\documentclass[conference]{IEEEtran}
\IEEEoverridecommandlockouts

\usepackage{cite}
\usepackage{amsmath,amssymb,amsfonts}
\usepackage{graphicx}
\usepackage{textcomp}
\usepackage{xcolor}
\usepackage[linesnumbered,ruled]{algorithm2e}
\usepackage{tabu,booktabs}
\usepackage{hyperref}
\hypersetup{
     colorlinks=true,
     linkcolor=red,
     filecolor=blue,
     citecolor=blue,      
     urlcolor=magenta,
     }

\usepackage{subcaption}

\def\BibTeX{{\rm B\kern-.05em{\sc i\kern-.025em b}\kern-.08em
    T\kern-.1667em\lower.7ex\hbox{E}\kern-.125emX}}
\begin{document}

\title{MiCRO: Near-Zero Cost Gradient Sparsification for Scaling and Accelerating Distributed DNN Training}

\author{\IEEEauthorblockN{Daegun Yoon and Sangyoon Oh}
\textit{Ajou University}\\
\{kljp, syoh\}@ajou.ac.kr}

\maketitle

\begin{abstract}
Gradient sparsification is a communication optimisation technique for scaling and accelerating distributed deep neural network (DNN) training. It reduces the increasing communication traffic for gradient aggregation. However, existing sparsifiers have poor scalability because of the high computational cost of gradient selection and/or increase in communication traffic. In particular, an increase in communication traffic is caused by gradient build-up and inappropriate threshold for gradient selection.

To address these challenges, we propose a novel gradient sparsification method called MiCRO. In MiCRO, the gradient vector is partitioned, and each partition is assigned to the corresponding worker. Each worker then selects gradients from its partition, and the aggregated gradients are free from gradient build-up. Moreover, MiCRO estimates the accurate threshold to maintain the communication traffic as per user requirement by minimising the compression ratio error. MiCRO enables near-zero cost gradient sparsification by solving existing problems that hinder the scalability and acceleration of distributed DNN training. In our extensive experiments, MiCRO outperformed state-of-the-art sparsifiers with an outstanding convergence rate.
\end{abstract}


\section{Introduction}\label{sec:1}
Over the past decade, overcoming the excessive communication traffic for gradient aggregation has been a major challenge to enhancing the distributed training performance of deep neural network (DNN) models. Gradient sparsification\cite{adacomp,learnedcomp,errorcompensatedx,natural,accordion,gtopk,lagssgd,omgssgd,oktopk} is a widely-adopted solution for reducing the size of payloads in communication between workers. Gradient sparsification aims to select only large gradients from the entire gradient vector, and the number of sparsified gradients is quantified by the density\footnote{We refer to the ratio of selected gradients ($k$) to all gradients ($n_g$) as density, which is defined as $d=\frac{k}{n_g}$.}. Therefore, gradient sparsification can alleviate the communication bottleneck when the communication bandwidth is insufficient to transmit all the gradients at every training iteration.

\begin{figure}[t]
    \centering
    \begin{subfigure}[t]{0.4\textwidth}
        \centering
        \includegraphics[width=1.0\linewidth]{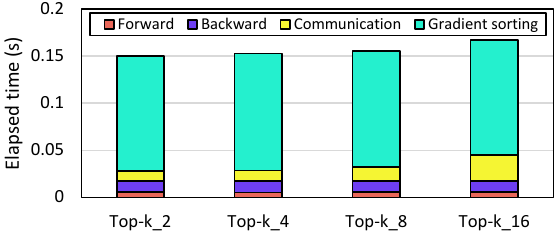}
        \caption{}
        \label{fig:1a}
    \end{subfigure}
    ~ 
    \begin{subfigure}[t]{0.42\textwidth}
        \centering
        \includegraphics[width=1.0\linewidth]{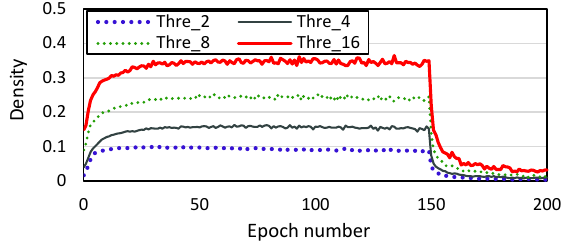}
        \caption{}
        \label{fig:1b}
    \end{subfigure}
    \caption{Challenges in scalable and accelerated gradient sparsification: (a) High computational cost due to gradient vector sorting in sorting-based sparsifiers; (b) high communication cost due to inappropriate threshold in threshold-based sparsifiers. Both types of sparsifiers cause gradient build-up. All experiments were conducted using $d=0.01$ and $n=\{2,4,8,16\}$ with ResNet-18 on CIFAR-10.}
    \label{fig:1}
\end{figure}

Gradient sparsification can be categorised into sorting- and threshold-based approaches. In sorting-based sparsifiers\cite{convproof01,scalecom}, all gradients are sorted, and the $k$ largest gradients are selected (top-k) for aggregation through communication. However, gradient vector sorting is an expensive operation because of its high computational complexity (e.g., $O({n}\log{k})$\cite{topkcomplexity}). Moreover, sorting operations cannot properly utilise the parallelism of streaming processors on GPUs\cite{drtopk}. Figure~\ref{fig:1a} shows the high computational cost for the gradient vector sorting of the Top-k sparsifier\cite{convproof01} based on the breakdown of the training time of one iteration. The computational cost remains constant and consumes a significant portion of the training time, regardless of the scale-out degree. Therefore, sorting-based sparsifiers are inadequate for accelerating distributed DNN training.

Threshold-based sparsifiers\cite{hardthreshold,sidco} select gradients using a conditional statement that indicates whether each gradient is larger than a threshold. Threshold-based sparsifiers are easier to parallelise and faster than sorting-based sparsifiers. Thus, threshold-based sparsifiers can significantly reduce the computational cost of gradient selection. However, threshold-based sparsifiers have difficulty effectively reducing communication traffic owing to inappropriate thresholds. Predicting a threshold that satisfies the density set by a user is challenging. Figure~\ref{fig:1b} shows the excessively high actual density of the hard-threshold sparsifier\cite{hardthreshold} where the user-set density was 0.01.

Additionally, most sparsifiers have difficulties scaling owing to gradient build-up\cite{scalecom}, which causes that the number of aggregated gradients in the communication becomes larger than the number of gradients selected by each worker. This is because a lot of gradients selected by each worker do not overlap with those of the other workers, although all workers have the same search range for gradient selection. Consequently, the density increases at most $n$ times the user-set density, where $n$ is the number of workers. As shown in Figure~\ref{fig:1}, the communication traffic increases as the number of workers increases because of the gradient build-up.

Therefore, the existing sparsifiers cannot effectively scale and accelerate distributed DNN training. Based on our observations, we address the following challenges:
\begin{itemize}
    \item \textbf{Gradient build-up}. This hinders the scalability of the distributed training because the communication traffic increases as the cluster scales out.
    \item \textbf{Inappropriate threshold}. Because the inappropriate threshold results in an extremely high actual density, it is difficult to reduce the communication traffic of the gradient aggregation and accelerate the distributed training.
    \item \textbf{Gradient selection cost}. Sorting the gradient vector for sparsification incurs a high computational cost. Because its high cost remains constant regardless of the scale-out of a cluster, it limits the acceleration of distributed DNN training.
\end{itemize}

In this study, we propose MiCRO\footnote{MiCRO is an acronym for \textbf{mi}nimising \textbf{c}ompression \textbf{r}atio error \textbf{o}n-the-fly.} to address these challenges. MiCRO divides the tensor of the entire model equally into multiple partitions and assigns them exclusively to workers to reduce the search range for the gradient selection from $n_g$ to $\frac{n_g}{n}$. With this partitioning approach, MiCRO can not only reduce the computational complexity of gradient selection but also prevent gradient build-up because each worker selects gradients from exclusively assigned partition.

To reduce the computational cost of the gradient selection, MiCRO adopts threshold-based sparsification instead of sorting-based sparsification\cite{convproof01,scalecom}. Moreover, the gradient selection of MiCRO is faster than that of existing threshold-based sparsifiers\cite{hardthreshold,sidco} because the gradient vector partitioning of MiCRO reduces the computational complexity of the gradient selection from $O(n_g)$ to $O(\frac{n_g}{n})$. In addition, the model accuracy of MiCRO can be maintained at the same level as that of other sparsifiers\cite{convproof01,hardthreshold}. This is because the result of filtering elements (gradients) larger than the threshold in an array (gradient vector) is invariant, regardless of whether the array is partitioned or not.

Furthermore, MiCRO satisfies the communication traffic at the user-set level by estimating the threshold more accurately. MiCRO estimates the threshold by minimising the compression ratio error, which is defined as the difference between the actual and user-set densities. Therefore, MiCRO can maintain a low communication cost throughout the training period by estimating the accurate threshold and eliminating gradient build-up. By addressing these challenges, MiCRO enables near-zero cost gradient sparsification for scalable and accelerated distributed DNN training.

This study makes the following contributions:
\begin{itemize}
    \item \textbf{Exclusive gradient selection}. This eliminates the gradient build-up. In other words, communication efficiency can be improved because exclusive gradients are aggregated between workers. Moreover, computational cost can be reduced as the cluster scales out. This primarily contributes to the scalability of distributed DNN training.
    \item \textbf{Accurate threshold estimation}. This prevents an excessively high density due to an inappropriate threshold. In other words, the communication traffic can be maintained as low as the user-set value. This mainly contributes to the acceleration of distributed DNN training.
    \item \textbf{Multidimensional evaluation}. This study provides an extensive set of experiments and analyses for performance and efficiency comparisons between MiCRO and state-of-the-art sparsifiers.
\end{itemize}

The remainder of this paper is organised as follows. Section~\ref{sec:2} presents the preliminaries of the study. Section~\ref{sec:3} clarifies the limitations of the state-of-the-art gradient sparsification methods. Section~\ref{sec:4} proposes MiCRO, which is designed to address the challenges stated in this study. Section~\ref{sec:5} verifies our contributions through thorough empirical comparisons between MiCRO and state-of-the-art gradient sparsifiers. Finally, Section~\ref{sec:6} concludes the paper.

\begin{figure*}[t]
    \centering
    \begin{subfigure}[t]{0.321\textwidth}
        \centering
        \includegraphics[width=1.0\linewidth]{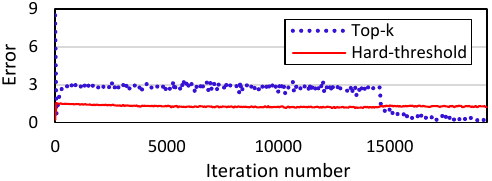}
        \caption{ResNet-18 on CIFAR-10 ($d=0.01$)}
        \label{fig:2a}
    \end{subfigure}
    ~ 
    \begin{subfigure}[t]{0.322\textwidth}
        \centering
        \includegraphics[width=1.0\linewidth]{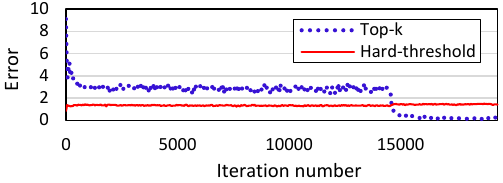}
        \caption{GoogLeNet on CIFAR-10 ($d=0.01$)}
        \label{fig:2b}
    \end{subfigure}
    ~ 
    \begin{subfigure}[t]{0.317\textwidth}
        \centering
        \includegraphics[width=1.0\linewidth]{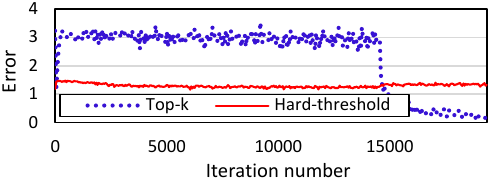}
        \caption{SENet-18 on CIFAR-10 ($d=0.01$)}
        \label{fig:2c}
    \end{subfigure}
    \caption{Error minimisation performance of sparsifiers on 16 GPUs. The Y-axis indicates the error, which is the average of local errors of workers.}
    \label{fig:2}
\end{figure*}

\begin{table*}
    \centering
    \caption{Strengths and weaknesses of state-of-the-art gradient sparsifiers and the proposed MiCRO.}
    \label{tab:1}
    \tabulinesep = 1mm
    \begin{tabu} to \linewidth {X[1.1,l,m]X[0.6,c,m]X[0.8,c,m]X[0.8,c,m]X[0.7,c,m]X[0.5,c,m]X[0.8,c,m]X[0.8,c,m]}
        \toprule
        Sparsifier & Gradient build-up & Unpredictable high density & Hyperparameter tuning & Model fidelity loss & Worker idling & Gradient selection cost & Additional overhead \\
        \midrule
        Top-k\cite{convproof01} & Yes & Yes & No & No & No & Very high & No \\
        CLT-k\cite{scalecom} & No & No & No & Yes & Yes & Very high & No \\
        Hard-threshold\cite{hardthreshold} & Yes & Yes & Yes & No & No & Very low & No \\
        SIDCo\cite{sidco} & Yes & Yes & No & No & No & Very low & Very high \\
        MiCRO & No & No & No & No & No & Near-zero  & No \\
        \bottomrule
    \end{tabu}
\end{table*}

\section{Preliminaries}\label{sec:2}
Gradient sparsification is a type of lossy algorithm because most of the computed gradients are discarded at every iteration. In terms of computational efficiency, discarding the majority of gradients is unproductive because backward propagation comprises a massive number of computational operations in DNNs. Moreover, discarded gradients cause a difference between sparsified and non-sparsified DNN models in distributed training. Thus, the fidelity loss of the sparsified model must be reduced to apply gradient sparsification to distributed training.

Error feedback\cite{seide2014} is an auxiliary method for sparsifiers to reduce the fidelity loss caused by discarded gradients. Instead of discarding unselected gradients, the error feedback locally accumulates them into the $n_g$-dimensional vector $e_{i,t}$, where $i$ and $t$ are the worker and iteration numbers, respectively. In other words, each element of $e_{i,t}$ represents the accumulated gradient of one parameter. When each gradient is selected, the accumulated value is initialised to zero because the gradient contributes to the model update. Hereafter, we refer to the L2-norm of $e_{i,t}$ as the local error denoted by ${\lVert}e_{i,t}{\rVert}$. Accordingly, the error\footnote{In this study, the terms `error' and `compression ratio error' are distinguished. The error indicates the difference between sparsified and non-sparsified models. The compression ratio error indicates the difference between actual and user-set densities.} at iteration $t$ is defined as follows:
\begin{equation}\label{equ:1}
    {\lVert}e_{t}{\rVert}=\frac{1}{n}\sum_{i=0}^{n-1}{{\lVert}e_{i,t}{\rVert}}.
\end{equation}

In other words, minimising ${\lVert}e_{t}{\rVert}$ results in a reduction of difference between sparsified and non-sparsified models. However, it is challenging to maintain a high model training performance while minimising ${\lVert}e_{t}{\rVert}$. To minimise ${\lVert}e_{t}{\rVert}$, the sparsifier should initialise a lot of accumulated gradients to zero in $e_{i,t}$. Because this implies a high density of sparsified gradients, the training performance slows down because of the huge communication traffic.

Figure~\ref{fig:2} shows the error variations in the two sparsifiers over the training iterations. The hard-threshold sparsifier selects all the gradients larger than the threshold, thus maintaining a consistent error level. However, the error of the hard-threshold sparsifier is much lower than that of Top-k in most iterations; thus, it is clear that the hard-threshold sparsifier can show a significantly higher density than Top-k. In other words, communication for the gradient aggregation of hard-threshold sparsifier is expensive. We verify how much runtime of hard-threshold sparsifier is occupied by the communication through experiments detailed in Section~\ref{sec:5}.

\section{Limitations of State-of-the-Art Methods}\label{sec:3}
In this section, we discuss the limitations of state-of-the-art gradient sparsifiers. Table~\ref{tab:1} lists the strengths and weaknesses of the state-of-the-art sparsifiers.

\subsection{Sorting-based sparsifiers}\label{sec:3.1}
In gradient sparsification, most computationally inefficient results are obtained from the gradient vector sorting phase of sorting-based sparsifiers, such as Top-k\cite{convproof01} and cyclic local top-k (CLT-k)\cite{scalecom}. The extremely high sorting cost is the main hindrance to the scalability and acceleration of distributed training.

In terms of communication efficiency, CLT-k maintains the user-set density by eliminating the gradient build-up of the Top-k. In CLT-k, a worker becomes the leader worker at each iteration and is delegated to determine all the gradient indices to be aggregated. Consequently, the number of aggregated gradients is the same as that of the selected gradients. However, the delegation policy for gradient selection has two side effects. First, most of the computing resources used by all other workers cannot be utilised during the gradient selection of the leader worker. Second, the model fidelity may be reduced because only one worker determines the gradient indices that contribute to the model update at each iteration. Therefore, Top-k and CLT-k exhibit limitations in terms of scalability and training performance.

\begin{figure*}
\centering
 \includegraphics[width=1.0\linewidth]{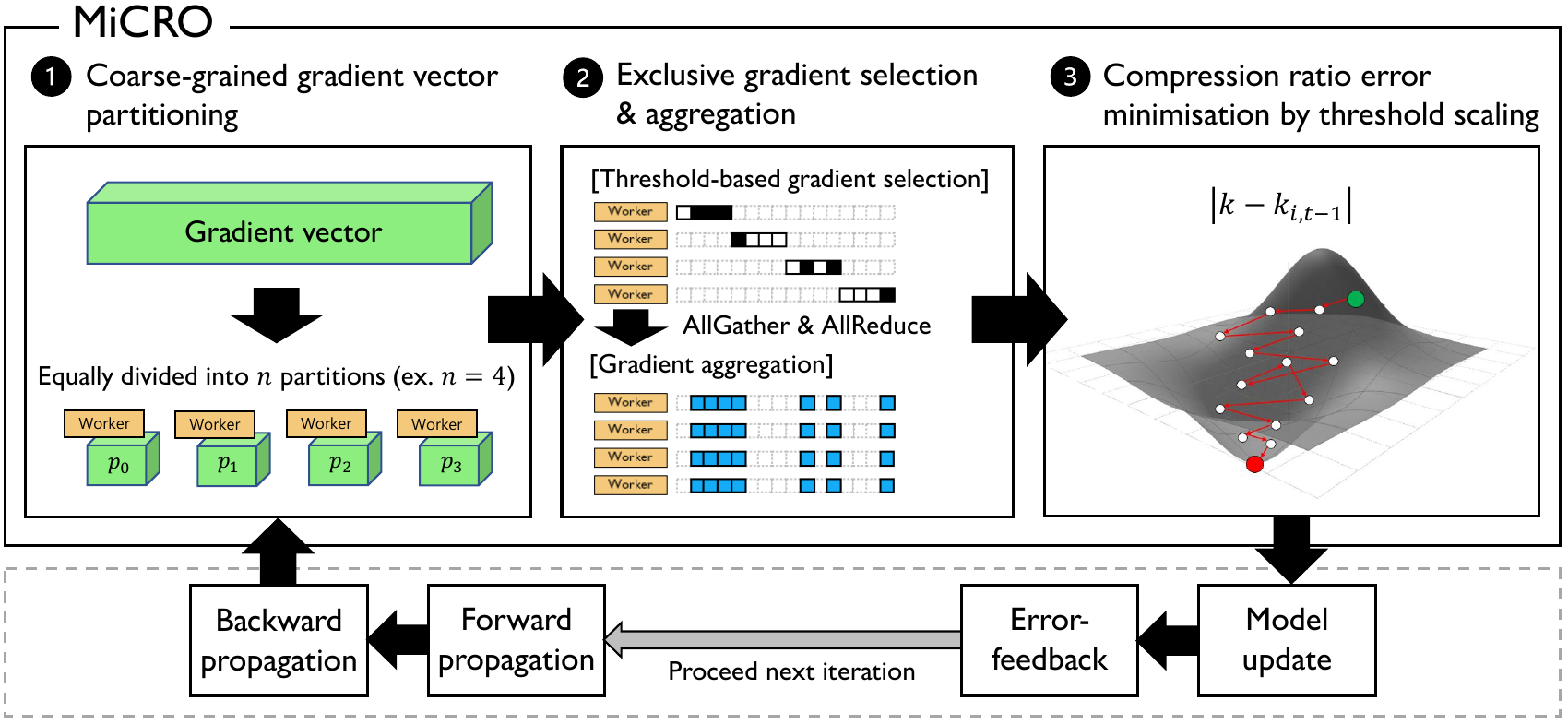}
\caption{Overview of MiCRO.}
\label{fig:3}
\end{figure*}

\subsection{Threshold-based sparsifiers}\label{sec:3.2}
The computational cost of gradient selection can be reduced considerably using threshold-based sparsifiers such as hard-threshold sparsifier\cite{hardthreshold} and SIDCo\cite{sidco}. However, they have limitations in terms of communication inefficiency. In addition to the gradient build-up problem, the actual density is excessively high because of inaccurate threshold estimation. This is mainly because their threshold estimation methods are insufficiently generalised to fit various types of models and datasets well. In particular, hard-threshold sparsifier may not estimate the appropriate threshold with untested training settings because its threshold should be defined before training begins and remain constant for every iteration. Therefore, hard-threshold sparsifier requires hyperparameter tuning for each model and dataset.

In contrast, SIDCo derives a threshold using a statistical model at each iteration. Because the threshold changes over the training iterations, the density can be adjusted more flexibly than with a hard-threshold sparsifier. However, SIDCo is based on several predefined statistical models, and threshold estimation by statistical models requires a high computational overhead. In summary, both hard-threshold sparsifier and SIDCo exhibit limitations in terms of efficient communication for scalable distributed training.

The limitations of state-of-the-art sparsifiers show that it is challenging to satisfy all the criteria listed in Table~\ref{tab:1}. We not only overcome the limitations of state-of-the-art methods using a novel sparsifier but also achieve it with near-zero cost.

\section{MiCRO Design}\label{sec:4}
We designed MiCRO as a threshold-based gradient sparsifier, in which each worker selects gradients from the exclusive partition of the entire gradient vector. Figure~\ref{fig:3} presents an overview of MiCRO. MiCRO comprises the following sequences: 1) coarse-grained gradient vector partitioning; 2) exclusive gradient selection by threshold; and 3) minimisation of compression ratio error by threshold scaling. These processes begin after backward propagation at each iteration. The following subsections present a detailed discussion of each process of MiCRO.

\subsection{Coarse-grained gradient vector partitioning}\label{sec:4.1}
MiCRO equally divides the entire gradient vector into $n$ partitions and assigns each partition to the corresponding worker. By gradient vector partitioning, each worker can obtain a search range that is exclusive to that particular worker. The partition assignment is not fixed for every iteration. At the beginning of each iteration, the partitioned vectors are assigned to the workers in a cyclic order. That is, each worker has the opportunity to search for gradients in the entire gradient vector in every $n$ iterations, and it also has the chance to select its local gradients at every iteration.

As shown in Figure~\ref{fig:3}, the entire gradient vector is partitioned in a coarse-grained manner. Each partitioned vector has a subsequent range of gradient indices. We designed this coarse-grained gradient vector partitioning method by considering the GPU memory access pattern. Because each thread group of the GPU (i.e., warp in CUDA\cite{cudadoc}) accesses the global memory simultaneously, the elements that the threads want to access should be in a cache line to utilise parallelism. Therefore, coarse-grained partitioning enables efficient GPU global memory access, unlike fine-grained partitioning, such as interleaved partitioning, which severely degrades performance owing to memory divergence\cite{memorydiv}.

\subsection{Exclusive gradient selection}\label{sec:4.2}
To prevent gradient build-up, a sparsifier should provide workers with a non-overlapping search space. Because the gradient build-up results from overlapping search spaces between workers, MiCRO restricts the search space of each worker to one partitioned vector, which is divided by coarse-grained gradient vector partitioning. Accordingly, each worker in the MiCRO can select gradients in its exclusively partitioned gradient vector, and gradient build-up never occurs. As shown in Figure~\ref{fig:3}, exclusive gradient selection enables nonoverlapping selected gradients.

Notably, MiCRO prevents loss of model fidelity, unlike CLT-k\cite{scalecom}. In CLT-k, each worker has the chance to select its local gradients once every $n$ iterations. However, workers have no selection authority for the remaining $n-1$ iterations. In other words, the locally computed and accumulated gradients of each worker will become stale. By contrast, MiCRO does not suffer from model fidelity loss because all workers can participate in the model update at every iteration. Moreover, the threshold-based gradient selection of MiCRO prevents model fidelity loss because all gradients are inspected to determine whether each of them is larger than the threshold. In other words, the significance of selecting the largest $\sum_{i=0}^{n-1}{k_{i,t}}$ gradients is maintained.

In addition, exclusive gradient selection reduces the computational complexity of threshold-based gradient selection from $O(n_g)$ to $O(\frac{n_g}{n})$. Therefore, scalability is enhanced because of the reduction in computational cost as the number of workers increases.

\subsection{Compression ratio error minimisation by threshold scaling}\label{sec:4.3}
To prevent a high density caused by an inappropriate threshold, a sparsifier should estimate the accurate threshold to achieve the user-set density. MiCRO focuses on minimising the compression ratio error at each iteration. Let $k$ and $k_{i,t}$ be the number of gradients that should be selected and the number of gradients selected by worker $i$ at iteration $t$, respectively. If $|k-k_{i,t}|$ is close to zero, it implies that the threshold is appropriate to satisfy the user-set value. Thus, minimising $|k-k_{i,t}|$ is crucial for adjusting the communication traffic of gradient aggregation.

As shown in Figure~\ref{fig:3}, $k_{i,0}$ may be far from the user set $k$ because it is difficult to predict the initial threshold $\delta_{0}$ accurately at iteration $0$. To minimise $|k-k_{i,t}|$, MiCRO adjusts the threshold at each iteration by inspecting whether the number of selected gradients are larger than $k$. This threshold scaling has two advantages over statistical threshold estimation\cite{sidco}. First, it is robust to varying training settings such as models, datasets, and learning parameters because only the compression ratio error is considered when adjusting the threshold. Moreover, additional overhead does not occur because inspecting $|k-k_{i,t}|$ and adjusting the threshold are performed by inspecting the condition statement and merely assigning an adjusted value to the threshold, respectively. Therefore, the threshold scaling of MiCRO is generally applicable to various training settings and is faster than statistical model-based threshold estimation.

\begin{algorithm}[t]
\SetAlgoVlined
\SetAlgoCaptionSeparator{}
\SetNlSty{}{}:{}
\PrintSemicolon
\SetKwInput{KwInput}{Input}
\KwInput{$G(\cdot)$: stochastic gradients\newline
Sparsify($\cdot$): threshold-based gradient sparsifier\newline
$n_g$: number of gradients in model
}
\For{worker $i$ ${\in}$ $[0,n-1]$ \textnormal{\textbf{in parallel}}}{
Initialise model $x_0$ ${\in}$ ${\mathbb{R}}^{n_g}$\;
Initialise local error $e_{i,0}$ ${\leftarrow}$ $0^{n_g}$\;
Initialise threshold $\delta_{0}$\;
\For{iteration $t$ ${\geq}$ $0$}{
${acc}_{i,t}$ ${\leftarrow}$ $e_{i,t}+{\eta}_{t}G_{i,t}(x_t)$\;
cycle ${\leftarrow}$ ($t$ $\%$ $n$ + $i$) $\%$ $n$\;
${part}_{i,t}$ ${\leftarrow}$ Partition(${acc}_{i,t}$, cycle)\;
${idx}_{i,t}$ ${\leftarrow}$ Sparsify(${part}_{i,t}$, $\delta_{t}$)\;
${idx}_t$ ${\leftarrow}$ AllGather(${idx}_{i,t}$)\;
$g_{i,t}$ ${\leftarrow}$ ${acc}_{i,t}[{idx}_t]$\;
$g_t$ ${\leftarrow}$ AllReduce($g_{i,t}$, SUM)\;
$\delta_{t+1}$ ${\leftarrow}$ Estimate($k$, $k_{i,t}$, $\delta_{t}$)\;
$x_{t+1}$ ${\leftarrow}$ $x_t-\frac{1}{n}g_t$\;
${acc}_{i,t}[{idx}_t]$ ${\leftarrow}$ $0$\;
$e_{i,t+1}$ ${\leftarrow}$ ${acc}_{i,t}$\;
}
}
\caption{Distributed SGD with MiCRO}
\label{alg:1}
\end{algorithm}

\subsection{Overall workflow of MiCRO}\label{sec:4.4}
Algorithm~\ref{alg:1} presents the pseudocode of the distributed SGD with gradient sparsification of MiCRO. In line 6, the gradients computed by backward propagation accumulate in local error. In line 7, the dedicated partition number is assigned to each worker in cyclic order. In line 8, the entire gradient vector is divided into $n$ partitions, and each partition is assigned to a dedicated worker. In line 9, each worker selects gradients based on the threshold in its exclusive partition and returns the indices of the selected gradients. According to the partition and selection policies, gradient build-up never occurs because the selected indices of each worker do not overlap with those of the other workers. From lines 10 to 12, the globally selected indices are collected and the averages of the selected gradients are computed. In line 13, the threshold of the next iteration is derived based on the compression ratio error minimisation. As the iterations proceed, the threshold approaches a value that satisfies the user-set density. In line 14, the model is updated using averaged gradients. In lines 15 and 16, the accumulated value of each selected gradient is initialised to zero, and those of the unselected gradients become the local error of the next iteration. 

\begin{table}
    \centering
    \caption{Description of each DNN application. $n$: number of workers, $B_l$: local batch size, $n_e$: number of epochs.}
    \label{tab:2}
    \tabulinesep = 1mm
    \begin{tabu} to \linewidth {X[1.3,l,m]X[3.1,l,m]X[1.7,l,m]X[0.1,l,m]X[0.1,r,m]X[0.3,r,m]}
        \toprule
        Model & Dataset & Density & $n$ & $B_l$ & $n_e$ \\
        \midrule
        ResNet-18 & \{CIFAR-10, CIFAR-100\} & \{0.01, 0.001\} & 16 & 32 & 200\\
        GoogLeNet & \{CIFAR-10, CIFAR-100\} & \{0.01, 0.001\} & 16 & 32 & 200\\
        SENet-18 & \{CIFAR-10, CIFAR-100\} & \{0.01, 0.001\} & 16 & 32 & 200\\
        \bottomrule
    \end{tabu}
\end{table}

\begin{figure*}[t]
    \centering
    \begin{subfigure}[t]{0.321\textwidth}
        \centering
        \includegraphics[width=1.0\linewidth]{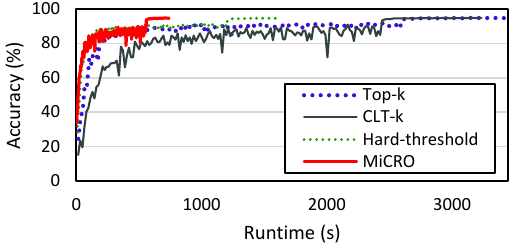}
        \caption{ResNet-18 on CIFAR-10 ($d=0.01$)}
        \label{fig:4a}
    \end{subfigure}
    ~ 
    \begin{subfigure}[t]{0.321\textwidth}
        \centering
        \includegraphics[width=1.0\linewidth]{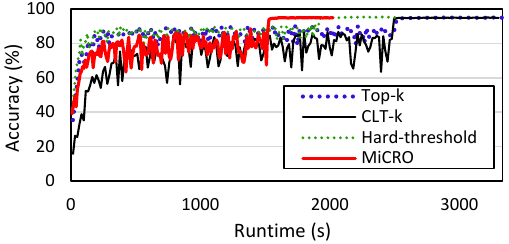}
        \caption{GoogLeNet on CIFAR-10 ($d=0.01$)}
        \label{fig:4b}
    \end{subfigure}
    ~ 
    \begin{subfigure}[t]{0.321\textwidth}
        \centering
        \includegraphics[width=1.0\linewidth]{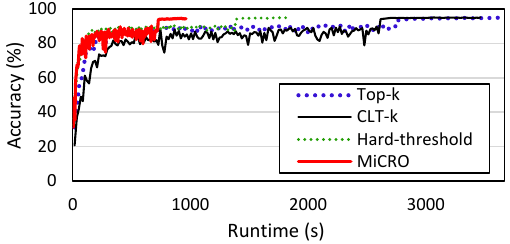}
        \caption{SENet-18 on CIFAR-10 ($d=0.01$)}
        \label{fig:4c}
    \end{subfigure}
    ~
    \begin{subfigure}[t]{0.321\textwidth}
        \centering
        \includegraphics[width=1.0\linewidth]{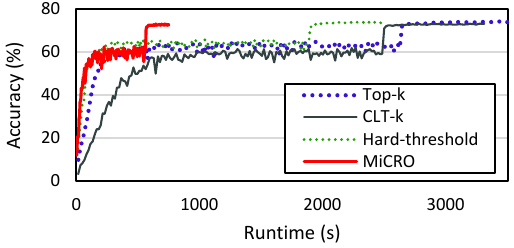}
        \caption{ResNet-18 on CIFAR-100 ($d=0.01$)}
        \label{fig:4d}
    \end{subfigure}
    ~ 
    \begin{subfigure}[t]{0.321\textwidth}
        \centering
        \includegraphics[width=1.0\linewidth]{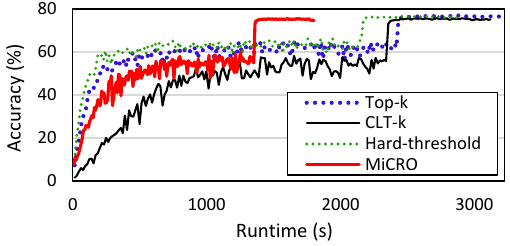}
        \caption{GoogLeNet on CIFAR-100 ($d=0.01$)}
        \label{fig:4e}
    \end{subfigure}
    ~ 
    \begin{subfigure}[t]{0.321\textwidth}
        \centering
        \includegraphics[width=1.0\linewidth]{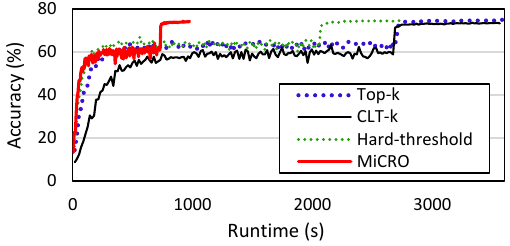}
        \caption{SENet-18 on CIFAR-100 ($d=0.01$)}
        \label{fig:4f}
    \end{subfigure}
    ~
    \begin{subfigure}[t]{0.321\textwidth}
        \centering
        \includegraphics[width=1.0\linewidth]{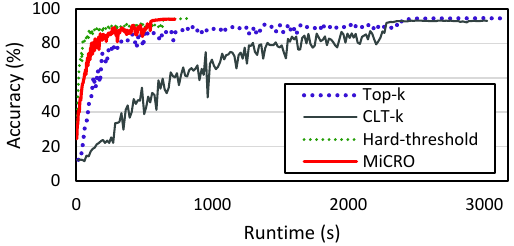}
        \caption{ResNet-18 on CIFAR-10 ($d=0.001$)}
        \label{fig:4g}
    \end{subfigure}
    ~ 
    \begin{subfigure}[t]{0.321\textwidth}
        \centering
        \includegraphics[width=1.0\linewidth]{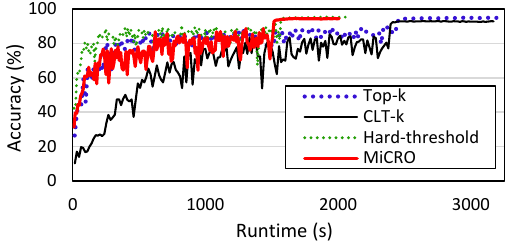}
        \caption{GoogLeNet on CIFAR-10 ($d=0.001$)}
        \label{fig:4h}
    \end{subfigure}
    ~ 
    \begin{subfigure}[t]{0.321\textwidth}
        \centering
        \includegraphics[width=1.0\linewidth]{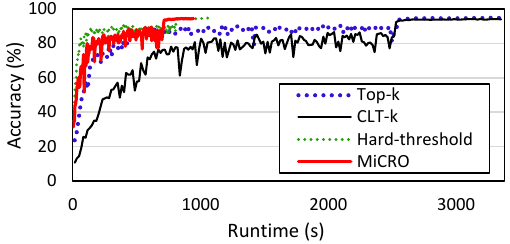}
        \caption{SENet-18 on CIFAR-10 ($d=0.001$)}
        \label{fig:4i}
    \end{subfigure}
    ~
    \begin{subfigure}[t]{0.321\textwidth}
        \centering
        \includegraphics[width=1.0\linewidth]{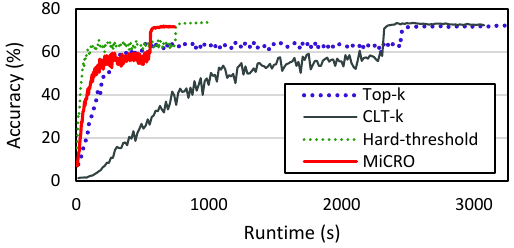}
        \caption{ResNet-18 on CIFAR-100 ($d=0.001$)}
        \label{fig:4l}
    \end{subfigure}
    ~ 
    \begin{subfigure}[t]{0.321\textwidth}
        \centering
        \includegraphics[width=1.0\linewidth]{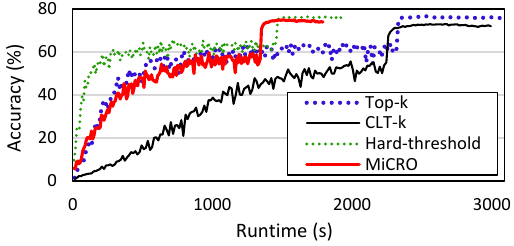}
        \caption{GoogLeNet on CIFAR-100 ($d=0.001$)}
        \label{fig:4m}
    \end{subfigure}
    ~ 
    \begin{subfigure}[t]{0.321\textwidth}
        \centering
        \includegraphics[width=1.0\linewidth]{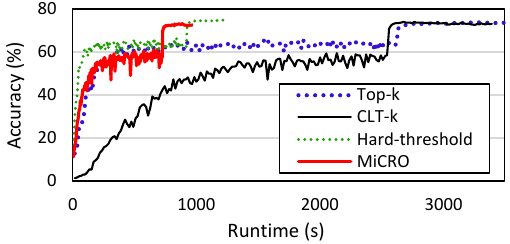}
        \caption{SENet-18 on CIFAR-100 ($d=0.001$)}
        \label{fig:4n}
    \end{subfigure}
    \caption{Convergence performance of sparsifiers on 16 GPUs. All experiments were conducted over 200 epochs.}
    \label{fig:4}
\end{figure*}

\section{Evaluation}\label{sec:5}
\subsection{Methodology}\label{sec:5.1}
\textbf{System configuration}. All the experiments were conducted on a cluster equipped with 16 GPUs. A cluster comprises two nodes, each with eight NVIDIA Tesla V100 GPUs with NVLink, two 16-core Intel Xeon Gold 6226R @ 2.90 GHz CPUs, and 384 GB DDR4 memory. For all the experiments, mpirun of OpenMPI 4.0.5\cite{openmpidoc} was used for multiprocess execution to automatically assign an exclusive rank to each worker. Each worker was run on one GPU with CUDA 10.1\cite{cudadoc}.

\textbf{Models and datasets}. We evaluated the performance of MiCRO and other sparsifiers (Top-k, CLT-k, and hard-threshold sparsifiers) on computer vision applications. For the DNN models, we used ResNet-18\cite{resnet}, GoogLeNet\cite{googlenet}, and SENet-18\cite{senet}. For datasets, CIFAR-10 and CIFAR-100\cite{cifar} were used. To conduct an extensive set of experiments and analyses, our evaluation comprised multidimensional training settings, as listed in Table~\ref{tab:2}. By changing one factor among the model, dataset, and density, the impact of each factor on the performance of each sparsifier can be identified.

\textbf{Implementation}. We implemented MiCRO and other approaches on top of the deep learning framework PyTorch 1.5\cite{torchdoc}. The distributed communication package PyTorch was used to implement the communication routine for the distributed training. Moreover, NCCL 2.4\cite{nccl} was adopted as a backend to support multi-GPU and multi-node communication primitives such as broadcast, all-gather, and all-reduce, which are optimised for NVIDIA GPUs and networking. To fairly compare the appropriateness of the thresholds between the MiCRO and the hard-threshold sparsifier, the initial threshold $\delta_{0}$ of the MiCRO was set to that of the hard-threshold sparsifier. The source code includes everything required to reproduce the results of this study, and is available publicly at \url{https://github.com/kljp/micro}.

\textbf{Metrics}. The metrics used for each type of performance evaluation are as follows:
\begin{itemize}
    \item Convergence performance: The test accuracy by runtime was measured to evaluate how fast each sparsifier attained the final accuracy in 200 epochs.
    \item Sparsification performance: The actual density was measured to evaluate whether each sparsifier could satisfy the user-set density.
    \item End-to-end training performance: The average wall clock time for one iteration was measured to evaluate the overall training speed of each sparsifier based on the breakdown of the training time.
\end{itemize}

\begin{figure*}[t]
    \centering
    \begin{subfigure}[t]{0.321\textwidth}
        \centering
        \includegraphics[width=1.0\linewidth]{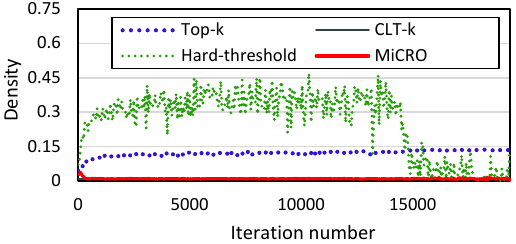}
        \caption{ResNet-18 on CIFAR-10 ($d=0.01$)}
        \label{fig:5a}
    \end{subfigure}
    ~ 
    \begin{subfigure}[t]{0.321\textwidth}
        \centering
        \includegraphics[width=1.0\linewidth]{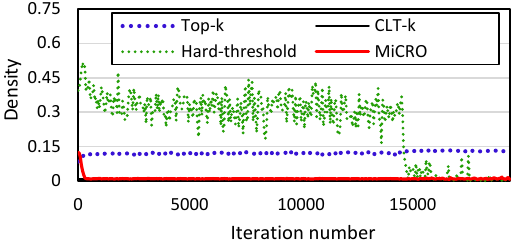}
        \caption{GoogLeNet on CIFAR-10 ($d=0.01$)}
        \label{fig:5b}
    \end{subfigure}
    ~ 
    \begin{subfigure}[t]{0.321\textwidth}
        \centering
        \includegraphics[width=1.0\linewidth]{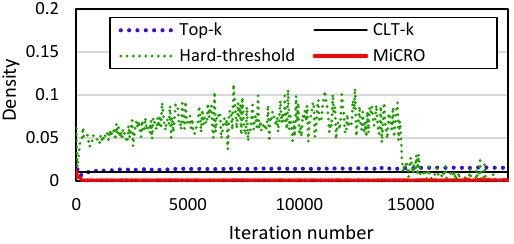}
        \caption{SENet-18 on CIFAR-10 ($d=0.01$)}
        \label{fig:5c}
    \end{subfigure}
    ~
    \begin{subfigure}[t]{0.321\textwidth}
        \centering
        \includegraphics[width=1.0\linewidth]{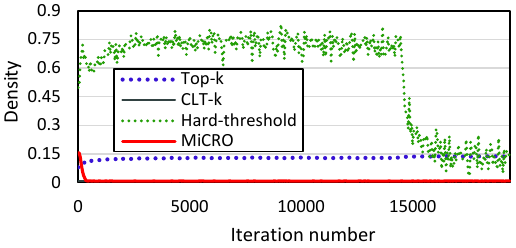}
        \caption{ResNet-18 on CIFAR-100 ($d=0.01$)}
        \label{fig:5d}
    \end{subfigure}
    ~ 
    \begin{subfigure}[t]{0.321\textwidth}
        \centering
        \includegraphics[width=1.0\linewidth]{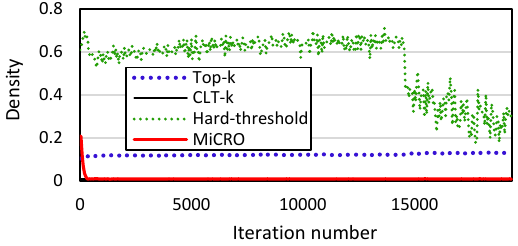}
        \caption{GoogLeNet on CIFAR-100 ($d=0.01$)}
        \label{fig:5e}
    \end{subfigure}
    ~ 
    \begin{subfigure}[t]{0.321\textwidth}
        \centering
        \includegraphics[width=1.0\linewidth]{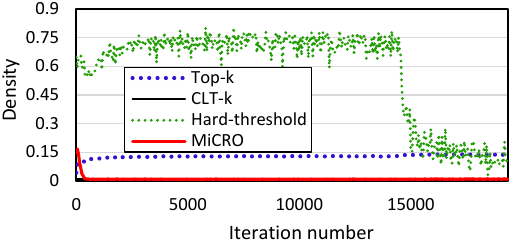}
        \caption{SENet-18 on CIFAR-100 ($d=0.01$)}
        \label{fig:5f}
    \end{subfigure}
    ~
    \begin{subfigure}[t]{0.321\textwidth}
        \centering
        \includegraphics[width=1.0\linewidth]{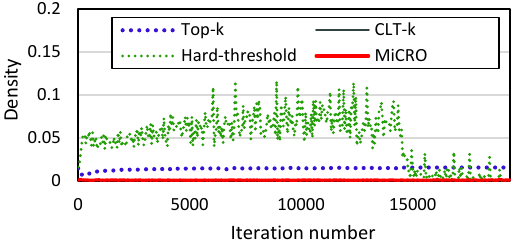}
        \caption{ResNet-18 on CIFAR-10 ($d=0.001$)}
        \label{fig:5g}
    \end{subfigure}
    ~ 
    \begin{subfigure}[t]{0.321\textwidth}
        \centering
        \includegraphics[width=1.0\linewidth]{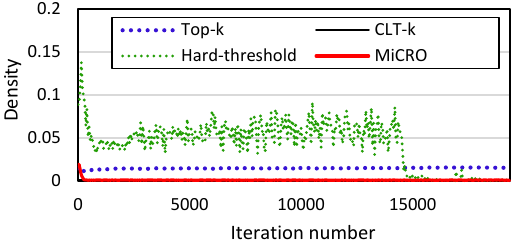}
        \caption{GoogLeNet on CIFAR-10 ($d=0.001$)}
        \label{fig:5h}
    \end{subfigure}
    ~ 
    \begin{subfigure}[t]{0.321\textwidth}
        \centering
        \includegraphics[width=1.0\linewidth]{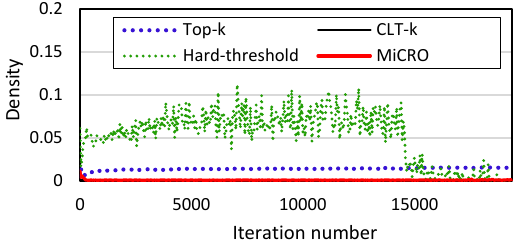}
        \caption{SENet-18 on CIFAR-10 ($d=0.001$)}
        \label{fig:5i}
    \end{subfigure}
    ~
    \begin{subfigure}[t]{0.321\textwidth}
        \centering
        \includegraphics[width=1.0\linewidth]{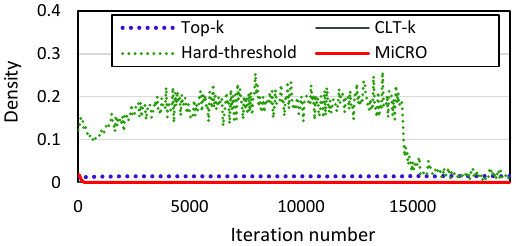}
        \caption{ResNet-18 on CIFAR-100 ($d=0.001$)}
        \label{fig:5l}
    \end{subfigure}
    ~ 
    \begin{subfigure}[t]{0.321\textwidth}
        \centering
        \includegraphics[width=1.0\linewidth]{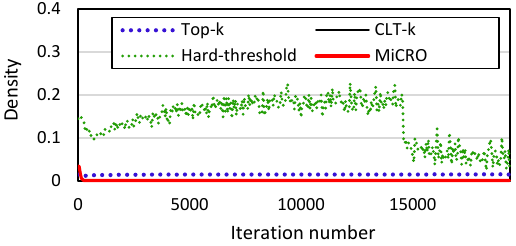}
        \caption{GoogLeNet on CIFAR-100 ($d=0.001$)}
        \label{fig:5m}
    \end{subfigure}
    ~ 
    \begin{subfigure}[t]{0.321\textwidth}
        \centering
        \includegraphics[width=1.0\linewidth]{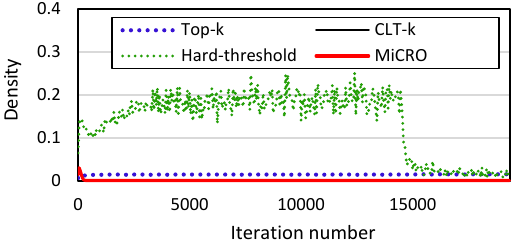}
        \caption{SENet-18 on CIFAR-100 ($d=0.001$)}
        \label{fig:5n}
    \end{subfigure}
    \caption{Sparsification performance of sparsifiers on 16 GPUs. The Y-axis indicates the actual density measured over training iterations.}
    \label{fig:5}
\end{figure*}

\subsection{Performance evaluation}\label{sec:5.2}
\textbf{Convergence performance}. Fig~\ref{fig:4} shows the convergence performance of each sparsifier under varying training settings for the model, dataset, and density. All experiments were completed at epoch 200, and each result showed the convergence rate by runtime. In every experiment, all sparsifiers attained a similar convergence point. However, the convergence rates of the sparsifiers differ because of their computational or communication efficiencies. In all cases, MiCRO shows the fastest convergence rate among the sparsifiers because its sparsification cost is almost zero owing to the threshold-based gradient selection and the elimination of gradient build-up.

However, CLT-k shows the slowest convergence rate among the sparsifiers because the training speed is limited by gradient sorting and the model fidelity is reduced by the delegated gradient selection policy. Top-k requires a long training time, similar to that of CLT-k. However, the convergence rate of Top-k was faster than that of CLT-k. This is because Top-k entails a gradient build-up, which makes Top-k select a lot more gradients than CLT-k (i.e., at most $n$ times the user-set density). Although the hard-threshold sparsifier has no computational cost for gradient sorting, its convergence rate is slower than that of MiCRO because it suffers from a large increase in communication traffic owing to the inappropriate threshold and gradient build-up.

\textbf{Sparsification performance}. Fig~\ref{fig:5} shows the sparsification performance of each sparsifier in multidimensional training settings. In every experiment, MiCRO exhibited the actual density close to the user-set density owing to the threshold based on compression ratio error minimisation. Moreover, gradient build-up does not occur because of the exclusive gradient selection with gradient vector partitioning.

However, the actual densities of the Top-k and hard-threshold sparsifiers were not close to the user-set densities because of gradient build-up. In particular, hard-threshold sparsifier exhibited the excessively high actual density owing to inappropriate threshold in every experiment. Despite the increasing gradient accumulation during training iterations, the hard-threshold sparsifier used only a fixed threshold. In other words, the actual density increased as the iterations proceeded. In every experiment, the density of the hard-threshold sparsifier dropped suddenly after iteration 14,600. This is because we set the learning rate decay at epoch 150. That is, the model almost converges after that epoch.

\begin{figure*}[t]
    \centering
    \begin{subfigure}[t]{0.321\textwidth}
        \centering
        \includegraphics[width=1.0\linewidth]{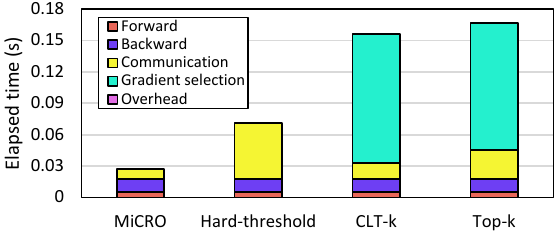}
        \caption{ResNet-18 on CIFAR-10 ($d=0.01$)}
        \label{fig:6a}
    \end{subfigure}
    ~ 
    \begin{subfigure}[t]{0.321\textwidth}
        \centering
        \includegraphics[width=1.0\linewidth]{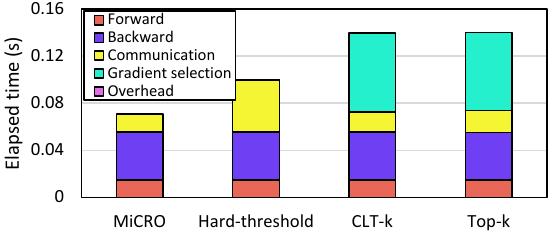}
        \caption{GoogLeNet on CIFAR-10 ($d=0.01$)}
        \label{fig:6b}
    \end{subfigure}
    ~ 
    \begin{subfigure}[t]{0.321\textwidth}
        \centering
        \includegraphics[width=1.0\linewidth]{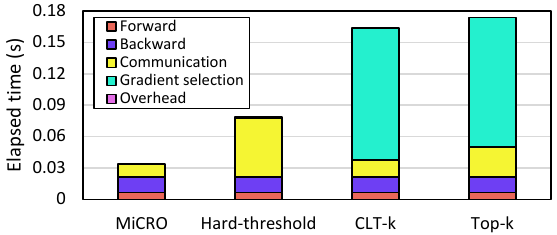}
        \caption{SENet-18 on CIFAR-10 ($d=0.01$)}
        \label{fig:6c}
    \end{subfigure}
    ~
    \begin{subfigure}[t]{0.321\textwidth}
        \centering
        \includegraphics[width=1.0\linewidth]{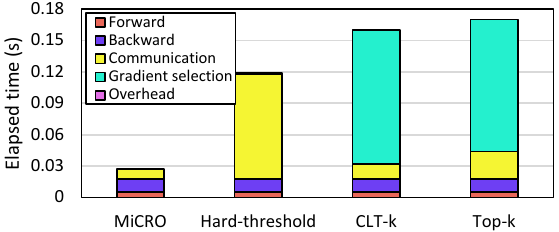}
        \caption{ResNet-18 on CIFAR-100 ($d=0.01$)}
        \label{fig:6d}
    \end{subfigure}
    ~ 
    \begin{subfigure}[t]{0.321\textwidth}
        \centering
        \includegraphics[width=1.0\linewidth]{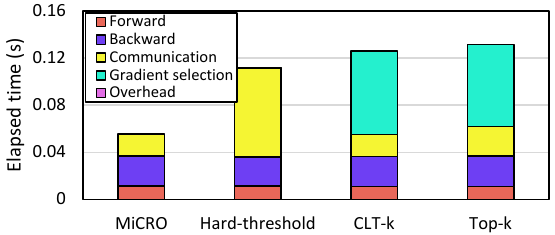}
        \caption{GoogLeNet on CIFAR-100 ($d=0.01$)}
        \label{fig:6e}
    \end{subfigure}
    ~ 
    \begin{subfigure}[t]{0.321\textwidth}
        \centering
        \includegraphics[width=1.0\linewidth]{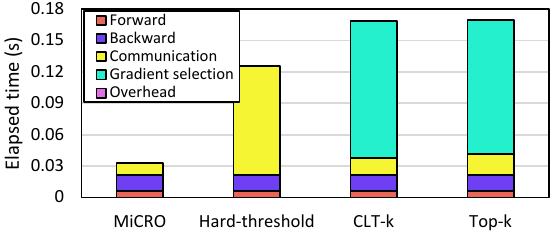}
        \caption{SENet-18 on CIFAR-100 ($d=0.01$)}
        \label{fig:6f}
    \end{subfigure}
    ~
    \begin{subfigure}[t]{0.321\textwidth}
        \centering
        \includegraphics[width=1.0\linewidth]{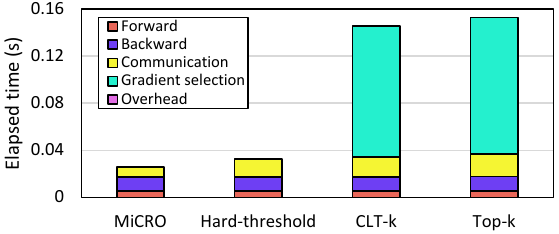}
        \caption{ResNet-18 on CIFAR-10 ($d=0.001$)}
        \label{fig:6g}
    \end{subfigure}
    ~ 
    \begin{subfigure}[t]{0.321\textwidth}
        \centering
        \includegraphics[width=1.0\linewidth]{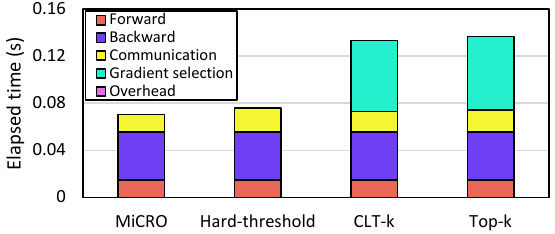}
        \caption{GoogLeNet on CIFAR-10 ($d=0.001$)}
        \label{fig:6h}
    \end{subfigure}
    ~ 
    \begin{subfigure}[t]{0.321\textwidth}
        \centering
        \includegraphics[width=1.0\linewidth]{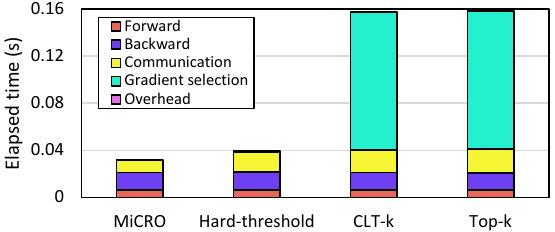}
        \caption{SENet-18 on CIFAR-10 ($d=0.001$)}
        \label{fig:6i}
    \end{subfigure}
    ~
    \begin{subfigure}[t]{0.321\textwidth}
        \centering
        \includegraphics[width=1.0\linewidth]{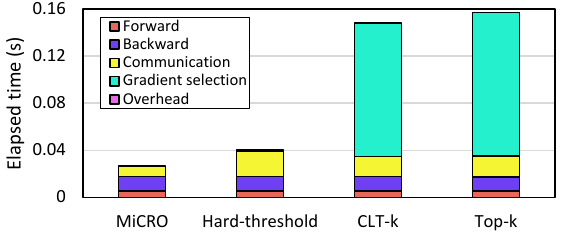}
        \caption{ResNet-18 on CIFAR-100 ($d=0.001$)}
        \label{fig:6l}
    \end{subfigure}
    ~ 
    \begin{subfigure}[t]{0.321\textwidth}
        \centering
        \includegraphics[width=1.0\linewidth]{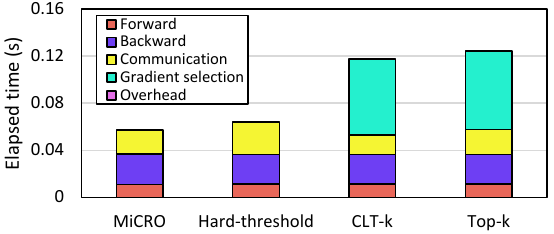}
        \caption{GoogLeNet on CIFAR-100 ($d=0.001$)}
        \label{fig:6m}
    \end{subfigure}
    ~ 
    \begin{subfigure}[t]{0.321\textwidth}
        \centering
        \includegraphics[width=1.0\linewidth]{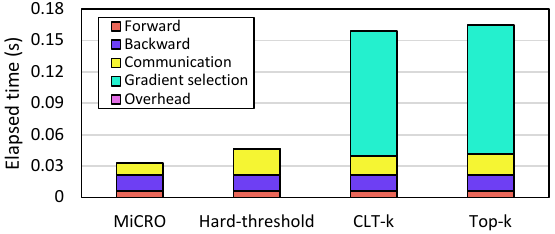}
        \caption{SENet-18 on CIFAR-100 ($d=0.001$)}
        \label{fig:6n}
    \end{subfigure}
    \caption{Training time breakdown of sparsifiers on 16 GPUs. The training time is the average wall-clock time for one iteration.}
    \label{fig:6}
\end{figure*}

\textbf{End-to-end training performance}. Fig~\ref{fig:6} shows the breakdown of the training time for one iteration. For each sparsifier, the wall-clock time of one iteration was measured by the slowest worker, and the average time was calculated across all iterations. The experiment was repeated using four different seeds for each sparsifier. Finally, the average wall-clock time shown in Figure~\ref{fig:6} was obtained from the average value of the four executions.

In Figure~\ref{fig:6}, the training time comprises the forward propagation, backward propagation, gradient selection, communication, and overhead times. The CLT-k and Top-k sparsifiers show much longer training times than the other sparsifiers owing to gradient vector sorting. Hard-threshold sparsifier exhibits long communication times in several cases owing to its inappropriate threshold and gradient build-up. These results indicate that a fixed threshold must be tuned for every training setting. In other words, hard-threshold sparsifier has limitations in general use in various training settings.

In contrast, MiCRO showed considerably faster training performance than the other sparsifiers in every experiment. This is because MiCRO reduces the computational cost to near-zero by using exclusive gradient selection with gradient vector partitioning and reduces the communication cost by eliminating gradient build-up and estimating the accurate threshold without overhead. As discussed in Section~\ref{sec:4.3}, the threshold estimation yields zero overhead because it only includes the condition statement for inspecting $|k-k_{i,t}|$ and the assignment of the adjusted value to a variable (i.e., the threshold). Moreover, the gradient vector partitioning of MiCRO does not yield any overhead because this process only determines the starting index of each partition. Therefore, when MiCRO is applied to distributed DNN training, the cost of gradient sparsification is near-zero, and this advantage contributes significantly to the scalability and acceleration of distributed training.

\begin{figure*}[t]
    \centering
    \begin{subfigure}[t]{0.321\textwidth}
        \centering
        \includegraphics[width=1.0\linewidth]{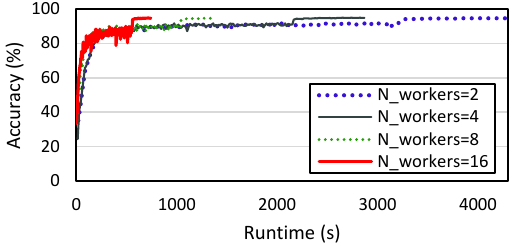}
        \caption{ResNet-18 on CIFAR-10 ($d=0.01$)}
        \label{fig:7a}
    \end{subfigure}
    ~ 
    \begin{subfigure}[t]{0.321\textwidth}
        \centering
        \includegraphics[width=1.0\linewidth]{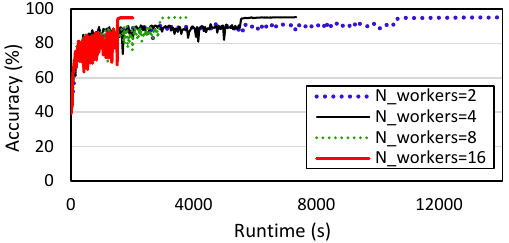}
        \caption{GoogLeNet on CIFAR-10 ($d=0.01$)}
        \label{fig:7b}
    \end{subfigure}
    ~ 
    \begin{subfigure}[t]{0.321\textwidth}
        \centering
        \includegraphics[width=1.0\linewidth]{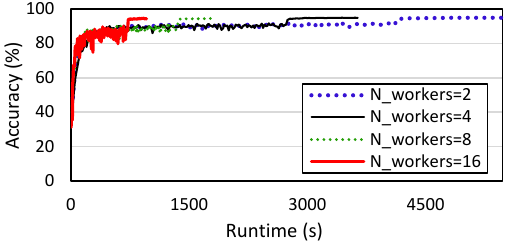}
        \caption{SENet-18 on CIFAR-10 ($d=0.01$)}
        \label{fig:7c}
    \end{subfigure}
    ~
    \begin{subfigure}[t]{0.321\textwidth}
        \centering
        \includegraphics[width=1.0\linewidth]{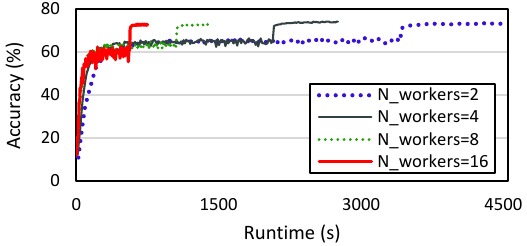}
        \caption{ResNet-18 on CIFAR-100 ($d=0.01$)}
        \label{fig:7d}
    \end{subfigure}
    ~ 
    \begin{subfigure}[t]{0.321\textwidth}
        \centering
        \includegraphics[width=1.0\linewidth]{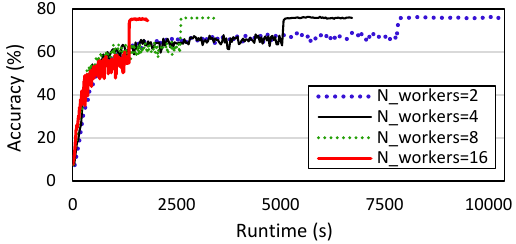}
        \caption{GoogLeNet on CIFAR-100 ($d=0.01$)}
        \label{fig:7e}
    \end{subfigure}
    ~ 
    \begin{subfigure}[t]{0.321\textwidth}
        \centering
        \includegraphics[width=1.0\linewidth]{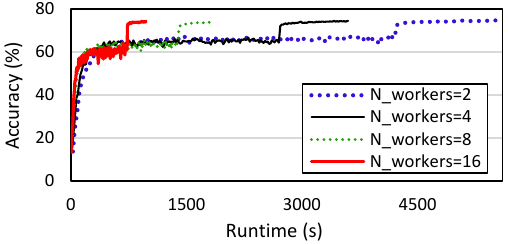}
        \caption{SENet-18 on CIFAR-100 ($d=0.01$)}
        \label{fig:7f}
    \end{subfigure}
    ~
    \begin{subfigure}[t]{0.321\textwidth}
        \centering
        \includegraphics[width=1.0\linewidth]{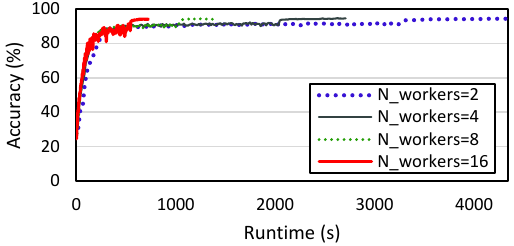}
        \caption{ResNet-18 on CIFAR-10 ($d=0.001$)}
        \label{fig:7g}
    \end{subfigure}
    ~ 
    \begin{subfigure}[t]{0.321\textwidth}
        \centering
        \includegraphics[width=1.0\linewidth]{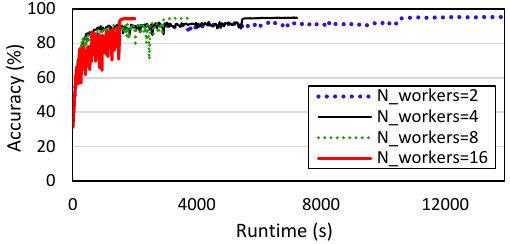}
        \caption{GoogLeNet on CIFAR-10 ($d=0.001$)}
        \label{fig:7h}
    \end{subfigure}
    ~ 
    \begin{subfigure}[t]{0.321\textwidth}
        \centering
        \includegraphics[width=1.0\linewidth]{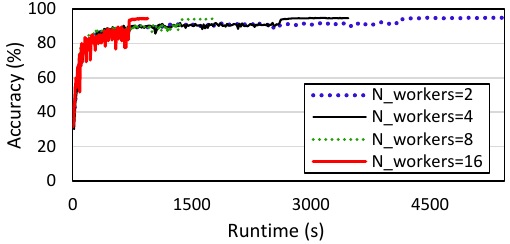}
        \caption{SENet-18 on CIFAR-10 ($d=0.001$)}
        \label{fig:7i}
    \end{subfigure}
    ~
    \begin{subfigure}[t]{0.321\textwidth}
        \centering
        \includegraphics[width=1.0\linewidth]{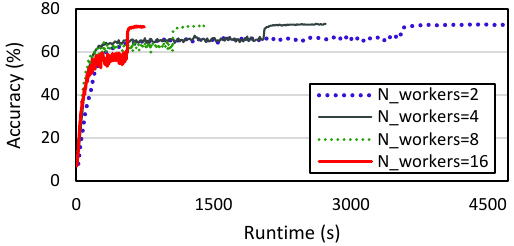}
        \caption{ResNet-18 on CIFAR-100 ($d=0.001$)}
        \label{fig:7l}
    \end{subfigure}
    ~ 
    \begin{subfigure}[t]{0.321\textwidth}
        \centering
        \includegraphics[width=1.0\linewidth]{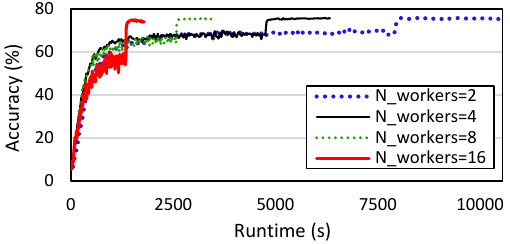}
        \caption{GoogLeNet on CIFAR-100 ($d=0.001$)}
        \label{fig:7m}
    \end{subfigure}
    ~ 
    \begin{subfigure}[t]{0.321\textwidth}
        \centering
        \includegraphics[width=1.0\linewidth]{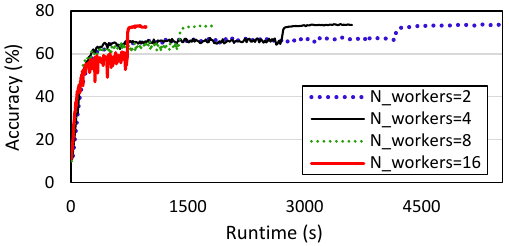}
        \caption{SENet-18 on CIFAR-100 ($d=0.001$)}
        \label{fig:7n}
    \end{subfigure}
    \caption{Convergence performance of MiCRO by scale-out. All experiments were conducted over 200 epochs.}
    \label{fig:7}
\end{figure*}

\begin{figure*}[t]
    \centering
    \begin{subfigure}[t]{0.321\textwidth}
        \centering
        \includegraphics[width=1.0\linewidth]{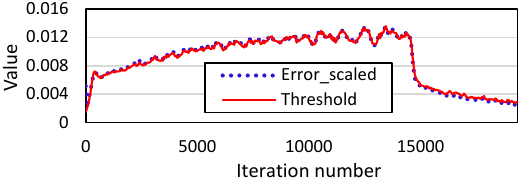}
        \caption{ResNet-18 on CIFAR-10 ($d=0.01$)}
        \label{fig:8a}
    \end{subfigure}
    ~ 
    \begin{subfigure}[t]{0.321\textwidth}
        \centering
        \includegraphics[width=1.0\linewidth]{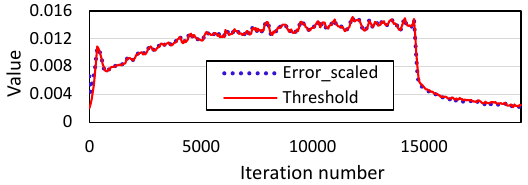}
        \caption{GoogLeNet on CIFAR-10 ($d=0.01$)}
        \label{fig:8b}
    \end{subfigure}
    ~ 
    \begin{subfigure}[t]{0.321\textwidth}
        \centering
        \includegraphics[width=1.0\linewidth]{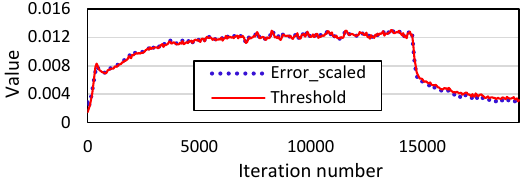}
        \caption{SENet-18 on CIFAR-10 ($d=0.01$)}
        \label{fig:8c}
    \end{subfigure}
    ~
    \begin{subfigure}[t]{0.321\textwidth}
        \centering
        \includegraphics[width=1.0\linewidth]{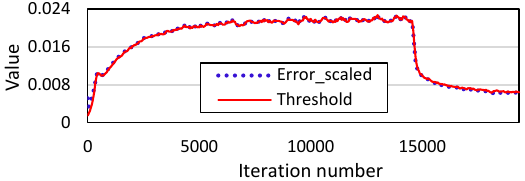}
        \caption{ResNet-18 on CIFAR-100 ($d=0.01$)}
        \label{fig:8d}
    \end{subfigure}
    ~ 
    \begin{subfigure}[t]{0.321\textwidth}
        \centering
        \includegraphics[width=1.0\linewidth]{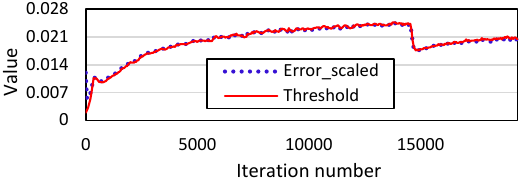}
        \caption{GoogLeNet on CIFAR-100 ($d=0.01$)}
        \label{fig:8e}
    \end{subfigure}
    ~ 
    \begin{subfigure}[t]{0.321\textwidth}
        \centering
        \includegraphics[width=1.0\linewidth]{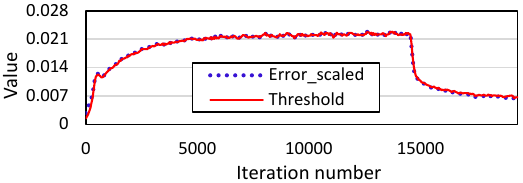}
        \caption{SENet-18 on CIFAR-100 ($d=0.01$)}
        \label{fig:8f}
    \end{subfigure}
    ~
    \begin{subfigure}[t]{0.321\textwidth}
        \centering
        \includegraphics[width=1.0\linewidth]{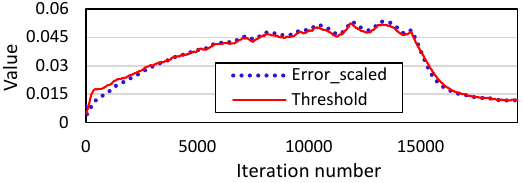}
        \caption{ResNet-18 on CIFAR-10 ($d=0.001$)}
        \label{fig:8g}
    \end{subfigure}
    ~ 
    \begin{subfigure}[t]{0.321\textwidth}
        \centering
        \includegraphics[width=1.0\linewidth]{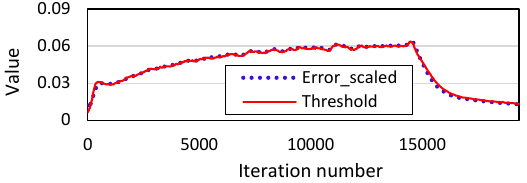}
        \caption{GoogLeNet on CIFAR-10 ($d=0.001$)}
        \label{fig:8h}
    \end{subfigure}
    ~ 
    \begin{subfigure}[t]{0.321\textwidth}
        \centering
        \includegraphics[width=1.0\linewidth]{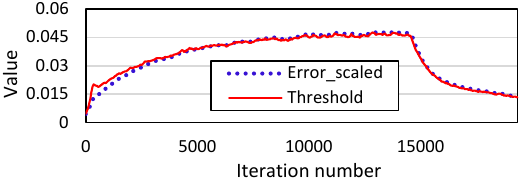}
        \caption{SENet-18 on CIFAR-10 ($d=0.001$)}
        \label{fig:8i}
    \end{subfigure}
    ~
    \begin{subfigure}[t]{0.321\textwidth}
        \centering
        \includegraphics[width=1.0\linewidth]{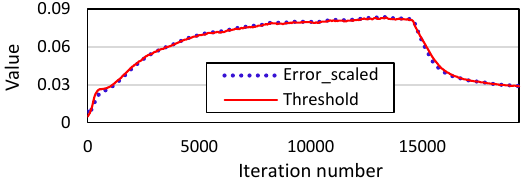}
        \caption{ResNet-18 on CIFAR-100 ($d=0.001$)}
        \label{fig:8l}
    \end{subfigure}
    ~ 
    \begin{subfigure}[t]{0.321\textwidth}
        \centering
        \includegraphics[width=1.0\linewidth]{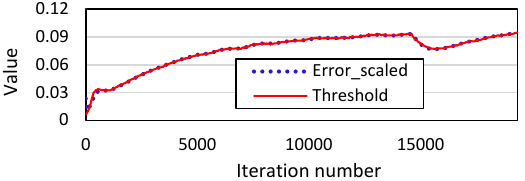}
        \caption{GoogLeNet on CIFAR-100 ($d=0.001$)}
        \label{fig:8m}
    \end{subfigure}
    ~ 
    \begin{subfigure}[t]{0.321\textwidth}
        \centering
        \includegraphics[width=1.0\linewidth]{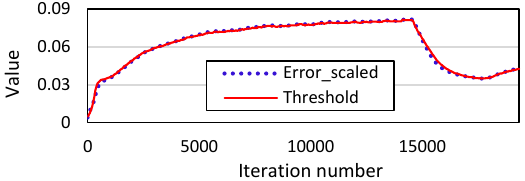}
        \caption{SENet-18 on CIFAR-100 ($d=0.001$)}
        \label{fig:8n}
    \end{subfigure}
    \caption{Threshold estimation performance of MiCRO on 16 GPUs.}
    \label{fig:8}
\end{figure*}

\subsection{Effectiveness evaluation of MiCRO}\label{sec:5.3}
\textbf{Scalability}. Fig~\ref{fig:7} shows the convergence performance of MiCRO by scale-out in our multidimensional training settings. MiCRO shows that every case consistently attains a similar convergence point, regardless of the number of workers. Moreover, the convergence rate was significantly accelerated by scale-out. These scalability and acceleration mainly result from communication cost reduction by eliminating gradient build-up and estimating the accurate threshold.

\textbf{Threshold estimation performance}. In this experiment, we evaluate the threshold estimation performance of the MiCRO. To maintain the accurate density, the threshold should be changed according to the error variation. That is, the threshold should increase when the error increases to prevent an increase in density. To identify whether MiCRO can satisfy this principle, we plotted the threshold and error trends. As the magnitude of the error is much larger than the threshold, we scaled the magnitude of the error to fit within a range of threshold changes. Thus, we multiplied the error of each iteration by the scaling factor, defined as the ratio of $\sum_{j=0}^{T-1}{\delta_{j}}$ to $\sum_{j=0}^{T-1}{{\lVert}e_{j}{\rVert}}$, where $T$ is the number of iterations.

Fig~\ref{fig:8} presents the threshold estimation performance of MiCRO in our multidimensional training setting. In every experiment, the threshold was properly changed according to the error trend. Through accurate threshold estimation based on compression ratio error minimisation, MiCRO can satisfy the communication traffic required by a user.

\section{Conclusion}\label{sec:6}
In this paper, we propose MiCRO, which partitions the gradient vector and assigns each partition to a corresponding worker. The design of MiCRO comprises coarse-grained gradient vector partitioning, exclusive gradient selection, and compression ratio error minimisation through threshold scaling. Using these components, MiCRO can achieve high performance in terms of convergence, sparsification, and threshold estimation. Consequently, it enables near-zero cost gradient sparsification by providing remarkable training efficiency owing to its reduced computational and communication costs. In our thorough empirical experiments, MiCRO outperformed state-of-the-art sparsifiers in terms of the scalability and acceleration of distributed DNN training.

\section*{Acknowledgment}
The authors would like to thank the anonymous reviewers for their insightful feedback. This work was jointly supported by the ITRC program (IITP-2023-2018-0-01431) of IITP, BK21 FOUR program (NRF5199991014091), and Basic Science Research Program (2021R1F1A1062779) of National Research Foundation of Korea.

\end{document}